\documentclass[conference]{IEEEtran}
\IEEEoverridecommandlockouts
\usepackage{cite}
\usepackage{amsmath,amssymb,amsfonts}
\usepackage{algorithmic}
\usepackage{graphicx}
\usepackage{textcomp}
\usepackage{xcolor}
\usepackage{multirow}
\usepackage{booktabs}
\usepackage{hyperref}
\usepackage{tablefootnote}
\usepackage[ruled,linesnumbered]{algorithm2e}
\usepackage{subfigure}
\usepackage{tablefootnote}
\usepackage{url}

\def\BibTeX{{\rm B\kern-.05em{\sc i\kern-.025em b}\kern-.08em
    T\kern-.1667em\lower.7ex\hbox{E}\kern-.125emX}}
\begin{document}

\title{DAGAD: Data Augmentation for Graph Anomaly Detection}

% \author{\IEEEauthorblockN{Fanzhen Liu\IEEEauthorrefmark{1}, Xiaoxiao Ma\IEEEauthorrefmark{1},
% Jia Wu\IEEEauthorrefmark{1},
% Jian Yang\IEEEauthorrefmark{1}, 
% Shan Xue\IEEEauthorrefmark{2},
% Amin Beheshti\IEEEauthorrefmark{1},
% Chuan Zhou\IEEEauthorrefmark{3}, \\
% Hao Peng\IEEEauthorrefmark{4},
% Quan Z. Sheng\IEEEauthorrefmark{1}, and
% Charu C. Aggarwal\IEEEauthorrefmark{5}}

\author{\IEEEauthorblockN{Fanzhen Liu\IEEEauthorrefmark{1}\textsuperscript{$\sharp$}\thanks{\noindent\textsuperscript{$\sharp$} Equal contribution.}, Xiaoxiao Ma\IEEEauthorrefmark{1}\textsuperscript{$\sharp$},
Jia Wu\IEEEauthorrefmark{1},
Jian Yang\IEEEauthorrefmark{1}, 
Shan Xue\IEEEauthorrefmark{2},
Amin Beheshti\IEEEauthorrefmark{1},
Chuan Zhou\IEEEauthorrefmark{3}, \\
Hao Peng\IEEEauthorrefmark{4},
Quan Z. Sheng\IEEEauthorrefmark{1}, and
Charu C. Aggarwal\IEEEauthorrefmark{5}}

\IEEEauthorblockA{\IEEEauthorrefmark{1}School of Computing, Macquarie University, Sydney, Australia}
\IEEEauthorblockA{\IEEEauthorrefmark{2}School of Computing and Information Technology, University of Wollongong, Wollongong, Australia}
\IEEEauthorblockA{\IEEEauthorrefmark{3}Academy of Mathematics and Systems Science, Chinese Academy of Sciences, Beijing, China}
\IEEEauthorblockA{\IEEEauthorrefmark{4}Beijing Advanced Innovation Center for Big Data and Brain Computing, Beihang University,
Beijing, China}
\IEEEauthorblockA{\IEEEauthorrefmark{5}IBM T. J. Watson Research Center, Yorktown, NY, USA}

{\{fanzhen.liu, xiaoxiao.ma2\}}@hdr.mq.edu.au, {\{jia.wu, jian.yang, amin.beheshti, michael.sheng\}}@mq.edu.au, \\ sxue@uow.edu.au, zhouchuan@amss.ac.cn, penghao@buaa.edu.cn, charu@us.ibm.com

}

% \and
% \and
% \and
% \and
% \and

\maketitle

\begin{abstract}
%background + motivation + proposed module
Graph anomaly detection in this paper aims to distinguish abnormal nodes that behave differently from the benign ones accounting for the majority of graph-structured instances. Receiving increasing attention from both academia and industry, yet existing research on this task still suffers from two critical issues when learning informative anomalous behavior from graph data. For one thing, anomalies are usually hard to capture because of their subtle abnormal behavior and the shortage of background knowledge about them, which causes severe anomalous sample scarcity. Meanwhile, the overwhelming majority of objects in real-world graphs are normal, bringing the class imbalance problem as well. To bridge the gaps, this paper devises a novel \underline{D}ata \underline{A}ugmentation-based \underline{G}raph \underline{A}nomaly \underline{D}etection (DAGAD) framework for attributed graphs, equipped with three specially designed modules: 1) an information fusion module employing graph neural network encoders to learn representations, 2) a graph data augmentation module that fertilizes the training set with generated samples, and 3) an imbalance-tailored learning module to discriminate the distributions of the minority (anomalous) and majority (normal) classes. A series of experiments on three datasets prove that DAGAD outperforms ten state-of-the-art baseline detectors concerning various mostly-used metrics, together with an extensive ablation study validating the strength of our proposed modules.

% could report improvement with xxx% 
\end{abstract}

\begin{IEEEkeywords}
Anomaly detection, graph mining, data augmentation, anomalous sample scarcity, class imbalance, graph neural networks, semi-supervised learning
\end{IEEEkeywords}

\section{Introduction}

Anomalies appear as objects that deviate from other reference members \cite{chandola2009anomaly, pang2021deep}. In various real-world scenarios, they could be fake news \cite{shu2017fake}, telecommunication fraudsters \cite{akoglu2015graph}, and spammers \cite{miller2014twitter}, which bring serious security and economic problems to our society.
% Anomaly detection is to dig out the objects that deviate from other reference ones, which has been widely applied to anti-money transfer, fake news detection
Benefiting from the power of graph modeling to characterize complicated interactions/relationships as connections among real-world objects \cite{chakrabarti2006graph, hamilton2017representation}, graph anomaly detection demonstrates its advantages in exposing anomalies by means of graph mining techniques \cite{akoglu2015graph, ma2021survey}, providing a comprehensive solution to dealing with complex graph-structured data. In this way, real-world anomalies can be depicted as - anomalous nodes representing single objects like fraudsters \cite{zhang2021fraudre}, anomalous edges denoting interactions like illegal transactions \cite{Weber2019anti}, and abnormal subgraphs revealing groups of interconnected malevolent objects, such as fraud groups \cite{liu2022eriskcom}. This work concentrates on detecting anomalous nodes that appear most frequently in real scenarios.

\begin{figure}[t]
    \centerline{\includegraphics[width=0.50\textwidth]{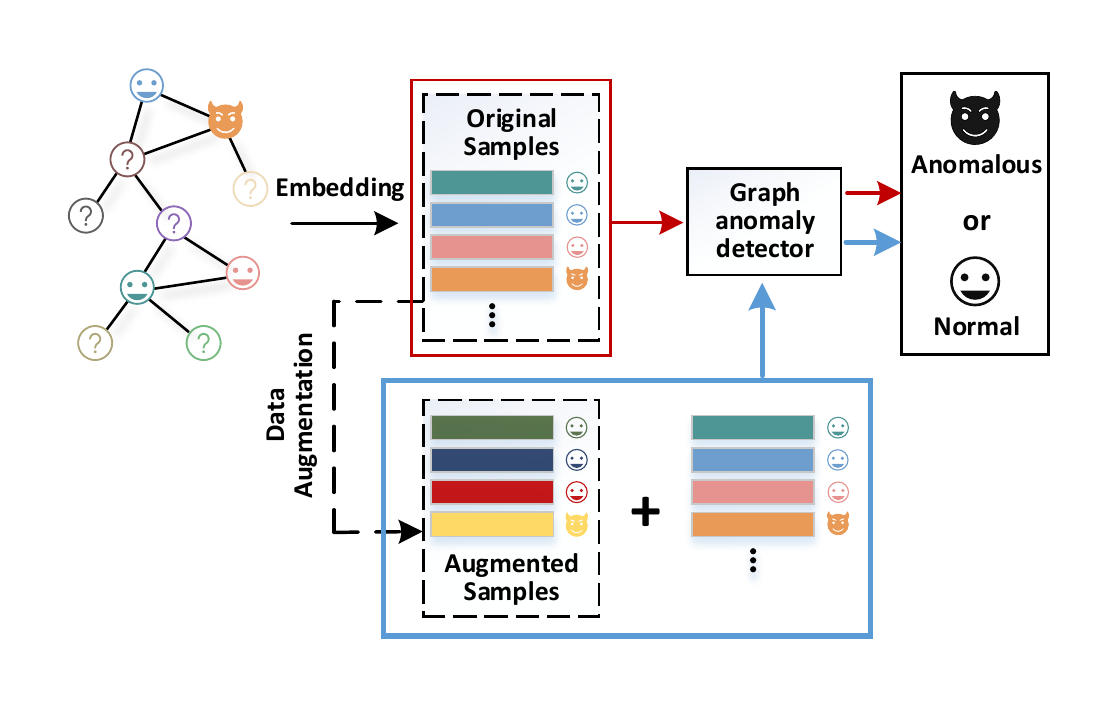}}
    \caption{A toy example of data augmentation for graph anomaly detection. With very few labeled samples, a graph anomaly detector (\textcolor{red}{$\rightarrow$}) that only exploits original data misidentifies some anomalies not easily exposed. Carefully augmenting training samples based on node embeddings/representations (\textcolor{cyan}{$\rightarrow$}) can complement information together with original samples to learn more effective graph anomaly detectors.}
    \label{pic:toy}
\end{figure}

Existing studies on graph anomaly detection have made efforts to discover anomalous objects dealing with graph topological information and rich features, but they are vulnerable to the intuitive nature of data regarding two issues, i.e., \textit{anomalous sample scarcity} and \textit{class imbalance}, to some extent. For one thing, real-world anomalies are not easy to observe. For instance, around 90\% of victims in e-commerce scenarios did not report through payment platforms like Alipay (www.alipay.com), so only a small number of anomalies can be captured \cite{liu2022eriskcom}. For another, anomalous objects are far less numerous than benign ones \cite{chalapathy2019deep}. As a result of this, graph anomaly detection is faced with the severely skewed distribution of anomalies versus benign nodes in quantity.

However, they fail to utilize the knowledge of even a limited number of anomalies, which to some extent sacrifices the capability to distinguish between normal and anomalous objects. 
% In practice, a small number of anomalies is available in most real-world cases.
Most of those unsupervised methods are built on autoencoders \cite{Hinton2006reducing} and rest on the assumption - that pursuing error minimization before and after data reconstruction is able to separate unusual items from the normal ones in a new low-dimensional feature space.
%Nevertheless, those approaches cannot cope with the impact of anomalies on the distribution of their normal neighbors. 
Others study graph anomaly detection assisted by labeled nodes \cite{Wang2019a,Ding2021few}, but they do not look into the class imbalance issue, incurring subpar anomaly detection performance.

In pursuit of better performance in anomaly detection, we develop a novel \textbf{D}ata \textbf{A}ugmentation-based \textbf{G}raph \textbf{A}nomaly \textbf{D}etection framework called DAGAD with three specially designed modules in tandem with each other to address the above two issues. DAGAD organizes these modules in a consolidated manner, summarized as follows:
1) an information fusion module encodes node attributes and graph topology information into low-dimensional vectors, a.k.a. node embeddings/representations, to represent fused features on nodes in a unified way;
2) a data augmentation module enriches the training set by generating additional training samples from original nodes based on their representations, which alleviates the suffering from \textit{anomalous sample scarcity}, as shown in Fig. \ref{pic:toy}; and
% 3) a mixed (mixture, blended) learning module attempts to derive the final node embeddings from the intermediate representations for anomalies and normal ones from the whole augmented dataset; and
3) an imbalance-tailored learning module comes up with a class-wise loss function to alleviate the class imbalance issue.
% workflow - a diagram?
Taking the advantage of graph neural networks (GNNs) in attributed graph learning \cite{zhou2020graph,Wu2021GNN}, DAGAD integrates the above modules into a GNN-aided learning framework to acquire an effective graph anomaly detector by extracting discriminative representations for anomalies and normal nodes. Most importantly, DAGAD is designed to exert maximum leverage on a very limited number of labeled data to distinguish anomalies.

%%%%%% related work? - unsupervised methods %%%%%%%%%%%%%%
% This workflow expects that normal items are to be clustered into a high-density region while anomalies are distributed outside of such region.

% To be specific, normal items are expected/guided to gather into a high-density region while anomalies are distributed outside such region. 
% Nevertheless, those approaches aggregating information from neighbors along edges cannot cope with the case where some anomalies have significant impacts on neighboring normal items by interactions. Hence, this leads to that these anomalies can not be easy to isolate from normal ones.

% 这里画个图举例说明一下？如果有真实的数据集。

% shed light on this issues regarding 

% anomaly scarcity指的很难发现；data imbalance指anomaly实际上发生的很少
\noindent
\textbf{Contributions.} This paper contributes to graph anomaly detection from the points as follows:
\begin{itemize}
    \item The investigated graph data augmentation technology generates additional samples derived from the original training set in the embedding space. Augmented samples together with original samples are leveraged by two classifiers in a complementary manner, to learn discriminative representations for the anomalous and normal classes.
    \item The representation-based data augmentation module in our framework provides a comprehensive solution to the scarcity of anomalous training samples in anomaly detection. This module is also extendable to other graph learning tasks that rely on learning features from a very limited number of labeled instances.
    \item A simple but effective imbalance-tailored learning module is employed to alleviate the suffering from class imbalance by utilizing a specially designed class-wise loss, which can be easily integrated into other semi-supervised graph anomaly detectors.
    \item Extensive experiments on three datasets as well as an ablation study prove DAGAD's superiority and the proposed modules' effectiveness under diverse evaluation criteria.
\end{itemize}

% \noindent
% \textbf{\color{red}Organization.}
% In the rest of this paper, we briefly review existing graph anomaly detection works, data augmentation studies, and class-imbalanced learning techniques especially on graph learning in Section~\ref{sec:relatedwork}, and provide preliminaries of this work in Section~\ref{prelim}. Section~\ref{fwk} details the devised modules of our proposed DAGAD framework, followed by extensive experiments and analysis proving the superiority and contributions of this work in Section~\ref{exp}. Lastly, we present the conclusions in Section~\ref{concl}.

\section{Related Work} \label{sec:relatedwork}

This paper focuses on the anomalous node detection problem, which aims to identify the nodes that significantly deviate from others in the graph. For completeness, we investigate recent studies on graph anomaly detection as well as data augmentation and class-imbalanced learning.

\subsection{Graph Anomaly Detection} \label{sec:relatedwork:gad}

To date, various graph anomaly detection studies have been conducted to identify potential anomalies (e.g., fraudsters and network intruders) in real-world networks \cite{akoglu2015graph, ma2021survey}. These studies explore the graph topology or non-structured node features from different perspectives for fusing the patterns of nodes and then identify anomalies that experience different patterns. Due to the advancement of deep graph data representation, especially graph neural networks \cite{zhou2020graph}, and their efficacy in graph analysis, uncovering graph anomalies with deep learning techniques has been extensively studied in contemporary works \cite{Luo2022ComGA, Bandyopadhyay2020outlier, wang2021ocgnn, Ding2019deep}.
Unlike conventional machine learning-based graph anomaly detection techniques that rely heavily on expert knowledge and human-recognized statistical features \cite{akoglu2010oddball, sharpnack2013near}, deep learning-based detectors deliver superior performance in wide applications ranging from finance to network security.

Most deep learning-based graph techniques stem from the motivation to encode the rich graph data into high-level node representations \cite{Wu2021GNN}. Graph anomalies can then be identified in an unsupervised manner by assigning anomaly scores regarding the reconstruction loss introduced by each node \cite{li2017Radar, Ding2019deep, Bandyopadhyay2020outlier, fan2020anomalydae}, distance to the majority of nodes \cite{wang2021ocgnn}, or through semi-supervised/supervised learning \cite{liu2019geniepath, wang2019fdgars} to train deep classifiers. This line of research counts heavily on the informativeness of node representations, and advanced graph neural network models such as GCN \cite{kipf2017gcn}, GAT \cite{velivckovic2018graph}, and GraphSAGE \cite{hamilton2017inductive}, are therefore widely adopted for extracting node representations. However, existing works almost fail to fully capitalize on a very limited number of anomalies from the training set, and the majority of them follow an introduced assumption that anomalies can be manifest in reconstruction error in an unsupervised manner. Even though there are a few works under semi-supervised learning settings \cite{Wang2019a,Ding2021few}, it is difficult for them to effectively confront the challenges of the scarcity of labeled anomalies and class imbalance associated with anomaly detection. Further efforts to bridge these gaps are of great demand for better anomaly detection solutions.

\subsection{Data Augmentation}

Data augmentation aims at enhancing the quantity and/or size of training data by either slightly modifying original data or generating synthetic instances from original data \cite{Shorten2019survey}. 
It has been proved that fields ranging from natural language processing \cite{feng2021survey} to computer vision \cite{lee2021learning} benefit from the power of data augmentation. Hence, data augmentation can serve as an effective tool to alleviate the lack of anomalous samples.

For more complicated graph-structured data, researchers have designed various augmentation techniques from the perspective of either graph topology or node attributes. Edge manipulation by adding or dropping edges on the original graph structure is popular for node-level tasks \cite{Suresh2021adversarial, Zhao2021data, zhu2021graph}.
%
% mainly focus on graph data augmentation?
% augmentation on topology
Besides, graph sampling provides an alternative idea to generate augmented graph samples from the original graph topology, which can be derived from a target distribution considering augmentation strength and data diversity \cite{Park2021metropolis}.
% augmentation on node attributes
Apart from previous works manipulating the graph structure, a recent work pursues better performance by employing adversarial perturbations to augment node features during training \cite{kong2022robust}. Also, data augmentation has been extended to graph-level tasks \cite{You2020graph, Qiu2020gcc, you2021graph}. However, performing manipulation on either original graph topology or node attributes could lead to inferior performance of models. 
Therefore, our work takes both the topology and node attribute into account and augments training samples by combining intermediate representations learned from two specially employed GNN encoders.

\subsection{Class-imbalanced Learning}

Due to the huge natural disparity in numbers of anomalies and normal objects, imbalanced class distribution constitutes an obstacle to graph anomaly detection \cite{zhang2021fraudre}. The class-imbalanced training data leads to deep learning models typically overfitting to the majority class because of their increased prior probability \cite{Johnson2019survey}. This means that heavily imbalanced data inflicts the “label bias” on the anomaly detector, where the majority class extremely alters the decision boundary \cite{yang2020rethink}.
Many attempts have been tried to overcome this challenge of class imbalance associated with deep learning \cite{Johnson2019survey}. Nevertheless, graph mining tasks not limited to graph anomaly detection still suffer from the lack of studies on this issue facing graph-structured data.
In class-imbalanced graph mining tasks, most existing graph learning techniques are prone to bias toward the majority class instances while under-train the minority classes. To achieve balanced learning between majority and minority node classes, \cite{Shi2020multi} presents an adversarial training strategy for cost-sensitive learning, while other data-level methods over-sample or generate nodes from minority classes \cite{Zhao2021GraphSMOTE,Qu2021ImGAGN}. 
However, for graph anomaly detection, it is not necessary to restrict synthetic anomalies to be generated from the minority class, as they only need to be different from the majority class. Indeed, future efforts on class-imbalanced learning are desperately desired for graph anomaly detection. 

% the objects of the minority classes are misclassified more often than those belonging to the majority classes. 

\section{Preliminaries}
\label{prelim}

\subsection{Definitions}

\textit{Attributed Graph.} $\mathcal{G=\left\{ V,E,A,X \right\}}$ represents an attributed graph with $n$ nodes, in which $\mathcal{V}$ and $\mathcal{E}$ denote the node set ${\left\{ v_i \right\}}_{i=1}^n$ and the edge set ${\left\{ e_{ij} \right\}}$ of $\mathcal{G}$, respectively. $e_{ij} = (v_i, v_j)$ denotes an edge connecting nodes $v_i$ and $v_j$, and $\mathcal{A} \in \left\{0,1\right\}^{n \times n}$ is an adjacency matrix storing $\mathcal{G}$'s topological structure, in which an entry $a_{ij}$ = 1 if $e_{ij} \in \mathcal{E}$; otherwise, $a_{ij}$ = 0. An attribute matrix $\mathcal{X} \in \mathbb{R}^{n \times k}$ stores the $k$-dimensional attribute $x_i$ of each node $v_i$.

\subsection{Problem Formulation}

Given an attributed graph $\mathcal{G}$,  graph anomaly detection can be resolved as a binary classification task. Specifically, each node $v_i$ should be identified to be either `normal' associated with a label of 0, or `anomalous' associated with a label of 1, i.e., $\mathcal{V} \rightarrow Y \in \left\{ 0,1 \right\}^n$. With observed labels for a limited number of nodes, this work aims to predict labels for other nodes.

\section{Proposed Framework}
\label{fwk}

\begin{figure*}
    \centering
    \includegraphics[width=\textwidth]{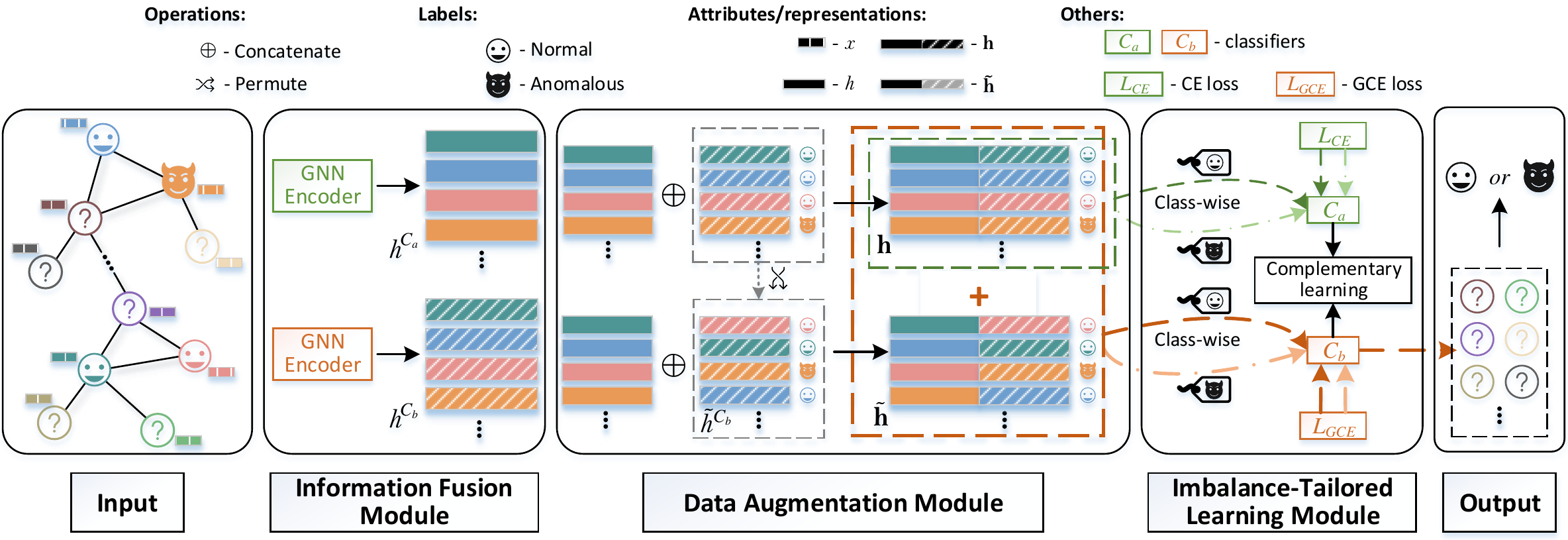}
    \caption{The framework of DAGAD. Taking an attributed graph as the input, DAGAD first employs two GNN encoders in fusing information on node attributes and the graph structure to extract node representations $h^{C_a}$ and $h^{C_b}$, respectively. Secondly, DAGAD concatenates $h^{C_a}$ and $h^{C_b}$ into $\mathbf{h}$ as representations for original nodes in the graph. Augmented training samples are generated by a concatenation of $h^{C_a}$ and permuted representations $\tilde{h}^{C_b}$ derived from $h^{C_b}$, denoted as $\tilde{\mathbf{h}}$. Labels of these augmented samples are assigned according to $\tilde{h}^{C_b}$'s labels. Lastly, the original samples represented by $\mathbf{h}$ are fed into classifier $C_a$ employing the class-wise CE loss and classifier $C_b$ employing the class-wise GCE loss, while the augmented samples $\tilde{\mathbf{h}}$ are fed into $C_b$ only. With the learning process going on in a complementary manner, each node will be identified as either anomalous or normal by $C_b$.}
    %The framework of DAGAD. Taking an attributed graph as the input, DAGAD first employs two GNN encoders in fusing information on node attributes and the graph structure to extract node representations $h^{C_a}$ and $h^{C_b}$, respectively. Secondly, the two representation components of the same node are combined into $\mathbf{h}$ by concatenation. In the same way, concatenating $h^{C_a}$ and $\tilde{h}^{C_b}$ which is permuted from $h^{C_b}$, DAGAD augments representations $\tilde{\mathbf{h}}$ for generated samples, and meanwhile, the labels assigned to generated samples are also according to the order of $\tilde{h}^{C_b}$ in $\tilde{\mathbf{h}}$. Thirdly, the original samples with representations are fed into classifier $C_a$ employing the class-wise CE loss and classifier $C_b$ employing the class-wise GCE loss, respectively, while the augmented samples are fed into $C_b$ only. With the learning process going on in a complementary manner, each node will be identified as either anomalous or normal by $C_b$.}
    \label{pic:framework}
\end{figure*}

As Fig.~\ref{pic:framework} shows, the proposed framework DAGAD is composed of three major modules working collaboratively.
The first information fusion module handles graph topology and node attributes and employs two GNN encoders to extract different intermediate representations of nodes, respectively. Then, the data augmentation module derives additional samples based on representations learned in the information fusion module to enrich the training set. Eventually, the imbalance-tailored learning module adopts class-wise losses to reveal the differences between anomalous and normal nodes with the class imbalance issue alleviated, and each node is labeled as normal or anomalous.

% 1) a feature/information fusion module encodes node attributes and graph topology information into low-dimensional vectors, a.k.a. node embeddings, to represent fused features on nodes in a unified way;
% 2) a data augmentation module fertilizes the training set by generating instances from original data in (both) quantity (and quality), which alleviates the suffering from \textit{data scarcity};
% 3) a mixed (mixture, blended) learning module attempts to derive the final node embeddings from the intermediate representations for anomalies and normal ones from the whole augmented dataset; and
% 4) a imbalance-tailored(-xxx-sensitive) classification module comes up with a class-wise(-independent) loss function to alleviate the imbalanced class distribution problem.

\subsection{Information Fusion Module}
\label{module1}

Performing anomaly detection on original graph-structured data is not easy, due to the fruitful node attributes and complicated graph topology. Instead, DAGAD tells the differences between anomalous and normal nodes based on their low-dimensional representations captured by encoders, i.e., $\phi: \left\{ \mathcal{A},\mathcal{X} \right\} \rightarrow H \in \mathbb{R}^{n \times d}$, where $H$ includes a node representation $h_i \in \mathbb{R}_d$ for each node $v_i$, satisfying $d \ll k$.

Taking advantage of the strength of GNN models in fusing graph topology and node attribute information into node representations \cite{Wu2021GNN}, DAGAD adopts a GNN model as an encoder to aggregate information from neighborhoods, which can be formulated as:
\begin{equation}
\label{GNN_encoder}
    h_i^l = f_{AGG}\Big(
    h_i^{l-1}, \big\{h_j^{l-1}: v_j \in \mathcal{N}(i)\big\}
    \Big),
\end{equation}
where $h_i^{l}$ denotes the representation associated with node $v_i$ in the $l$-th GNN layer, and other nodes standing in the neighborhood of node $v_i$ are collected in a set $\mathcal{N}(i)$. $f_{AGG}(\cdot)$ serves as an information aggregator (e.g., sum and mean) performed on node representations with taking in node attributes as initial node representations, i.e., $h_i^0 = x_i$.

\subsection{Data Augmentation Module} 
\label{module2}

\subsubsection{Augmentation on Representation}
Based on node representations learned by the information fusion module, anomaly detectors incur suboptimal performance as a result of exploiting limited information from original observed anomalous samples. Such a rarity of anomalies makes it difficult for graph anomaly detectors to distinguish anomalies and normal nodes clearly. Our data augmentation module assists anomaly detectors in addressing the anomalous sample scarcity issue from a new perspective.
Different from existing graph data augmentation techniques that augment either graph structures or node attributes, DAGAD's data augmentation module is aimed to generate samples based on node representations of original samples to enrich the knowledge of anomalies captured in the training set.

The whole learning process targets learning the boundary between anomalous and normal classes.
Specifically, DAGAD deploys two classifiers to capture discriminative features of anomalies and normal nodes from low-dimensional representations. Particularly, the performance of classifier $C_b$ will be enhanced with the help of another classifier $C_a$ which tries to extract the anomaly-related features as much as possible by overfitting to the anomalous class.

DAGAD augments samples involved in the training set by following the process:
First, we randomly permute the representations extracted by Eq. (\ref{GNN_encoder}): 
\begin{equation}
\label{permutate}
    \tilde{H} =  \text{PERMUTE}\big( \left\{h_{i}:v_i \in \mathcal{V}\right\} \big),
\end{equation}
where $\tilde{H}$ restores the permuted representations via $\text{PERMUTE}: h_i \rightarrow \tilde{h}_j \in \tilde{H}$.
Then, to make the above two classifiers work together, serving the fore-mentioned purpose, we concatenate the representations of two parts, each learned by a GNN encoder, that is,
\begin{equation}
\label{h_org}
   \mathbf{h}_i = \text{CONCAT}(h_i^{C_a},{h}_i^{C_b}),
\end{equation}
where $h_i^{C_a}$ and $h_i^{C_b}$ denote the GNN-encoded representations of node $v_i$ by classifiers $C_a$ and $C_b$, respectively. Analogous to this, the augmented samples will be obtained by a concatenation of $C_b$'s permuted representations and original ones of $C_a$'s by
\begin{equation}
\label{h_permuted}
   \tilde{\mathbf{h}}_i = \text{CONCAT}(h_i^{C_a},\tilde{h}_i^{C_b}),
\end{equation}
where $\tilde{h}_i^{C_b}$ is classifier $C_b$'s $i$-th representation after permutation. Furthermore, since $C_b$ will be trained using the augmented representations as well, the assignment of labels to augmented samples follows the permutation order, i.e, $\tilde{y}=\text{PERMUTE}(y)$.

\subsubsection{Complementary Learning} 

With augmentation performed on representations as a foundation, the training process proceeds with two above classifiers in a complementary fashion. 
In detail, each classifier employs a multiple-layer perceptron (MLP) $f^{\text{MLP}}(Z; \theta)$ (with 2 fully connected layers by default) \cite{Goodfellow2016deep} to obtain the final representations of nodes, where $Z$ can be their concatenated representations derived from the ones returned by GNN encoders and $\theta$ denotes the trainable parameter set. Taking input $\mathbf{h} \in \mathbb{R}^{2d}$ as an example here, each layer of the MLP can be formulated as:
\begin{equation}
    \mathbf{h}_i^{l} = \mathbf{h}_i^{l-1} W^\top + b
\end{equation}
where $W \in \mathbb{R}^{D \times 2d}$ and  $b \in \mathbb{R}^{D}$ are trainable weights and bias, getting a $D$-dimensional representation for each node. In this way, we can capture the final representations $\mathbf{h}^* \in \mathbb{R}^{2}$ for original samples and $\tilde{\mathbf{h}}^* \in \mathbb{R}^{2}$ for augmented samples.
Subsequently, a softmax function is employed to calculate the probability of a node being an anomaly or a normal instance by 
\begin{equation}
\label{Softmax}
    P(y|\mathbf{h^*}) = \text{Softmax}(\mathbf{h^*}).
\end{equation} 
Hence, each node will be assigned a label by
\begin{equation}
\label{label_predict}
    \mathbf{y} = \arg\max_{y} P(y|\mathbf{h^*}).
\end{equation}

\begin{figure}
    \centering
    \includegraphics[width=0.40\textwidth]{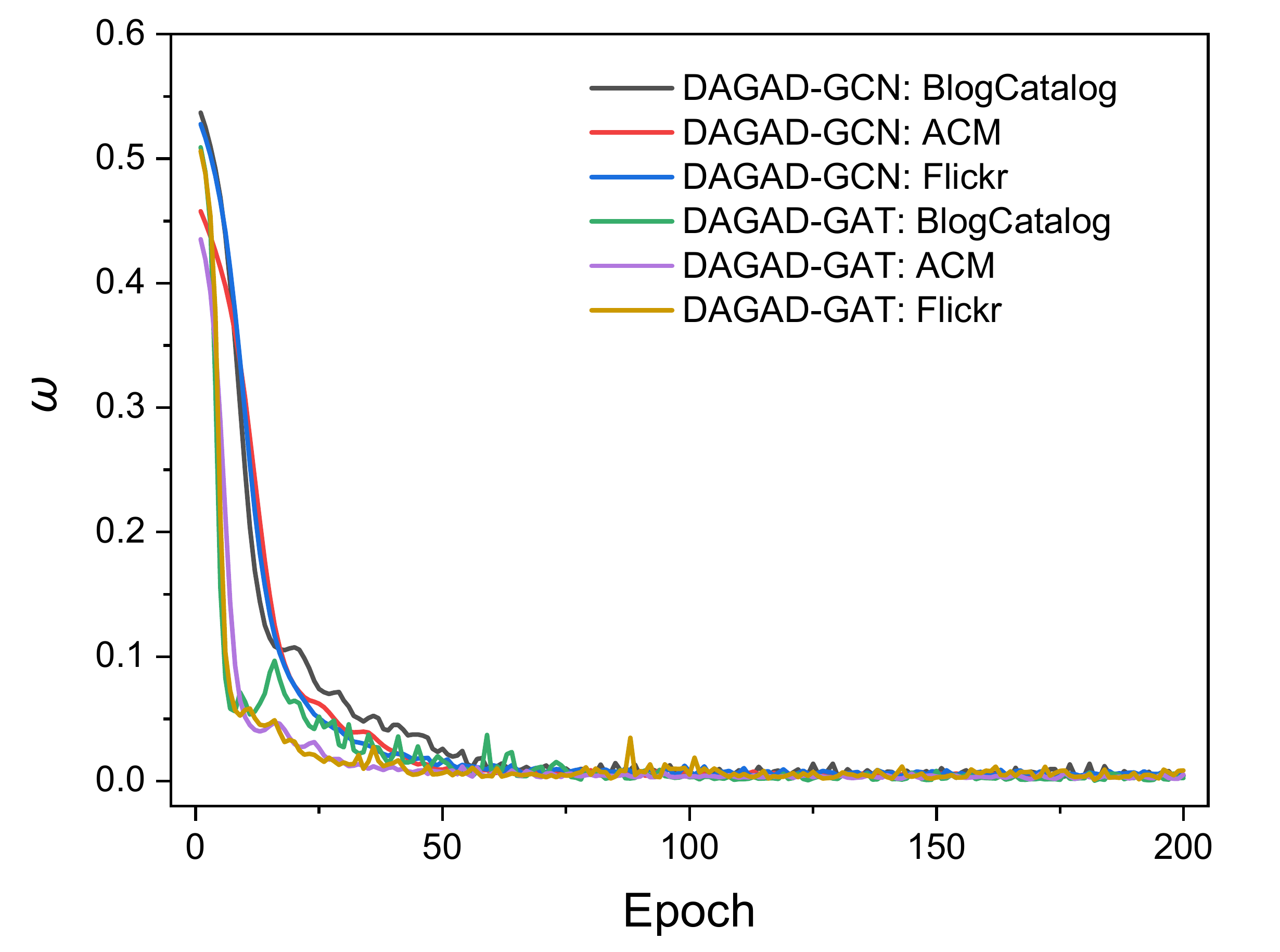}
    \caption{The values of $\omega(C_a,C_b,\mathbf{h^*},y)$ during the learning process of DAGAD on BlogCatalog, ACM, and Flickr.}
    \label{pic:weight_ce}
\end{figure}

As shown in Fig. \ref{pic:framework}, the learning process runs in a complementary manner, which optimizes the detection performance on the original training set as well as the augmented samples. For original training samples, we apply the cross entropy (CE) loss and the generalized cross entropy (GCE) loss \cite{zhang2018generalized} for $C_a$ and $C_b$, respectively. Classifier $C_a$ is expected to overfit to learn features most relevant to the anomalous class, which simultaneously helps $C_b$ capture the discriminative features between the anomalous class and the normal class, achieving a better detection result. Accordingly, the two parts are combined into 
\begin{equation}
\begin{aligned}
      \mathcal{L}_\text{org} &  =  \mathcal{L}_\text{org}^{C_a} + \mathcal{L}_\text{org}^{C_b} \\
      & =  \omega(C_a,C_b,\mathbf{h^*},y) \cdot \psi_{\text{CE}}\big(C_a(\mathbf{h^*}),y \big) \\
      & \ \ \ +  \psi_{\text{GCE}}\big(C_b(\mathbf{h^*}),y \big),
\end{aligned}
\end{equation}
where $\psi_{\text{CE}}$ and $\psi_{\text{GCE}}$ are the CE loss function and GCE loss function, respectively, and $\omega$ guarantees complementary information sharing between $C_a$ and $C_b$ \cite{nam2020learning}, calculated by  
\begin{equation}
\label{weight_ce}
    \omega(C_a,C_b,\mathbf{h^*},y)= \frac{\psi_{\text{CE}}\big(C_b(\mathbf{h^*}),y\big)}{\psi_{\text{CE}}\big(C_a(\mathbf{h^*}),y\big)+\psi_{\text{CE}}\big(C_b(\mathbf{h^*}),y\big)}.
\end{equation}
The decrease of $\omega(C_a,C_b,\mathbf{h^*},y)$ in value, as shown in Fig.~\ref{pic:weight_ce},  will restrict $C_a$ to overfit to the anomalous class during the learning process.

For another thing, $C_b$ also involves the augmented samples in the GCE loss by maximizing the objective:
\begin{equation}
    \mathcal{L}_\text{aug}^{C_b} = \psi_{\text{GCE}}\big(C_b(\tilde{\mathbf{h}}^*), \tilde{y} \big).
\end{equation}
Having all parts together, the overall loss function of DAGAD is formulated as:
\begin{equation}
\label{overall_loss}
    \mathcal{L} = \alpha \cdot \mathcal{L}_\text{org}^{C_a} + \mathcal{L}_\text{org}^{C_b} + \beta \cdot \mathcal{L}_\text{aug}^{C_b}.
\end{equation}
To enable that $h^{C_a}$ and $h^{C_b}$ are learned mainly under the guidance of $C_a$ and $C_b$, the loss from $C_b$ is not back-propagated to the encoder learning $h_i^{C_a}$, and vice versa.

%%%%the version in code %%%%
% \begin{equation}
%     L = (1 + \alpha) \cdot \omega(H) \cdot L_{C_i} + L_{C_b} + \alpha \cdot \beta \cdot L_{\text{comp}}
% \end{equation}

\subsection{Imbalance-tailored Learning Module}
\label{module3}

Suffering from imbalanced training data, graph learning-based anomaly detectors are always inclined to bias toward the normal class while under-training the anomalous class. DAGAD resolves this issue by coming up with an imbalance-tailored learning module. To balance the contributions from the two classes to the learning process, we develop a class-wise loss function based on Eq. (\ref{overall_loss}). Instead of treating every training sample equally, as shown in the standard instance-wise CE loss for classification:
\begin{equation}
    \psi_{\text{CE}} = - \frac{1}{|\mathcal{V}_{train}|} \sum_i^{|\mathcal{V}_{train}|}  p(v_i|y_i) \log p(v_i|y_i),
\end{equation}
where $p(v_i|y_i)$ is the probability of node $v_i$ belonging to its ground truth class $y_i$. We assign the same weight to both classes (i.e., anomalous vs. normal) by
\begin{equation}
\label{ce_new}
\begin{aligned}
    \widehat{\psi}_{\text{CE}} & =   \psi_{\text{CE}}\big(p(v_i | {y_i=1)}\big) + \psi_{\text{CE}}\big( p(v_j | {y_j=0})\big) \\
    & = - \frac{1}{|\mathcal{V}_{train}^{anm}|} \sum_i^{|\mathcal{V}_{train}^{anm}|} p(v_i|y_i) \log p(v_i|y_i) \\
    & \ \ \ - \frac{1}{|\mathcal{V}_{train}^{norm}|} \sum_j^{|\mathcal{V}_{train}^{norm}|} p(v_j|y_j) \log p(v_j|y_j),
\end{aligned}
\end{equation}
where $|\mathcal{V}_{train}^{anm}|$ and $|\mathcal{V}_{train}^{norm}|$ count anomalous and normal samples in the training set, respectively.
Similarly, a new class-wise GCE loss can be defined as:
\begin{equation}
\label{gce_new}
    \widehat{\psi}_{\text{GCE}}  =  \psi_{\text{GCE}}\big(p(v_i | {y_i=1)}\big) + \psi_{\text{GCE}}\big( p(v_j | {y_j=0})\big).
\end{equation}
On top of this, the overall loss as shown in Eq. (\ref{overall_loss}) is replaced by
\begin{equation}
\label{overall_loss_new}
    \widehat{\mathcal{L}}  = \alpha \cdot \widehat{\mathcal{L}}_\text{org}^{C_a} + \widehat{\mathcal{L}}_\text{org}^{C_b} + \beta \cdot \widehat{\mathcal{L}}_\text{aug}^{C_b},
\end{equation}
where $\{ \widehat{\mathcal{L}}_\text{org}^{C_a}, \widehat{\mathcal{L}}_\text{org}^{C_b}, \widehat{\mathcal{L}}_\text{aug}^{C_b} \}$ corresponds to  $\{ \mathcal{L}_\text{org}^{C_a}, \mathcal{L}_\text{org}^{C_b}, \mathcal{L}_\text{aug}^{C_b}\}$  updated by employing class-wise losses based on Eqs. \eqref{ce_new} and \eqref{gce_new}, instead of the instance-wise CE loss and GCE loss.

Algorithm \ref{DAGAD_algorithm} presents how the three major modules are organized in our framework DAGAD.

%%%% algorithm %%%
\begin{algorithm}[tbp]
\label{DAGAD_algorithm}
\caption{DAGAD}
\KwIn{attributed graph $\mathcal{G}$; training set with nodes $\mathcal{V}_{train}$ and labels $Y_{train}$; GNN encoders $\phi_a$ and $\phi_b$; classifiers $C_a$ and $C_b$; training iteration $T$}
\KwOut{predict labels $\mathbf{Y}$}
Initialize associated parameters \\
$t \gets 1$ \\
\While{$t \leq T$}{
    Extract $h^{C_a}$ and $h^{C_b}$ from $\phi_a$ and $\phi_b$ by Eq. \eqref{GNN_encoder} \\
    Obtain $\mathbf{h}$ by Eq. \eqref{h_org} and augmented $\tilde{\mathbf{h}}$ by Eq. \eqref{h_permuted} \\
    $\mathbf{h}^{*C_a} \gets \text{MLP}_{C_a}(\mathbf{h}$) \\
    $\mathbf{h}^{*C_b} \gets \text{MLP}_{C_b}(\mathbf{h})$ \\ 
    $\tilde{\mathbf{h}}^{*C_b} \gets \text{MLP}_{C_b}(\tilde{\mathbf{h}})$ \\
    Obtain $\mathbf{y_i}$ by Eqs. \eqref{Softmax} and \eqref{label_predict} \\
    Calculate $\widehat{\mathcal{L}}$ on the training set by Eq. \eqref{overall_loss_new} \\
    Update $\phi_a$, $\phi_b$, $C_a$, and $C_b$  by minimizing $\widehat{\mathcal{L}}$
}
\textbf{Return} $\mathbf{Y} = \left\{\mathbf{y}_i\right\}_{i=1}^{n}$
\end{algorithm}

\subsection{Computational Complexity}
\label{complexity}

During each training iteration, the computational cost of DAGAD is introduced by the forward- and backward-computation of the parameters in the GNN layers and MLPs. Taking the GCN as an example, the computation complexity will be approximately $\mathcal{O}(2nkd + 2dD)$ with $n$ node, $k$-dimensional node attributes, and $d$ and $D$ neurons in the MLPs. Additionally, the complexity of permutation-based augmentation approximates $\mathcal{O}(n_l)$ where $n_l < n$  counts labeled training samples. As in real cases, $D\leq d\ll k$, so the overall complexity of DAGAD approximates to $\mathcal{O}(nk^{2}+n)$.

\section{Experiments} 
\label{exp}

To conduct a performance evaluation of our devised framework, we implement extensive experiments by comparison with ten up-to-date graph anomaly detectors on three real-world attributed graph datasets. Besides, We perform ablation analysis to test the effectiveness of the specially designed functional modules. All these are expected to answer the following questions:
\begin{itemize}
    \item \textbf{Q1:} Is DAGAD superior to the up-to-date baseline approaches regarding anomaly detection performance?
    % \item \textbf{Q2:} Can DAGAD well discriminate anomalies from normal users in the learned embedding space?
    \item \textbf{Q2:} Does our data augmentation module facilitate the detection of anomalies?
    \item \textbf{Q3:} Does our imbalance-tailored learning module optimize anomaly detection performance?
    \item \textbf{Q4:} Is DAGAD sensitive to any major hyper-parameters?
\end{itemize}

\subsection{Experimental Setup}
\subsubsection{Datasets}
For fairness, we adopt the three most widely used real-world attributed graphs with injected anomalies in previous works \cite{Ding2019deep, ding2019interactive, fan2020anomalydae} to validate the effectiveness of our model. Table~\ref{tb:dataset_statistics} displays statistics of these three datasets\footnote{All datasets were downloaded on March 15, 2022 from \url{https://github.com/kaize0409/GCN_AnomalyDetection_pytorch/tree/main/data} \label{1}}. In our experiments, each dataset is divided into two parts which contain 20\% and 80\% of labeled nodes for training and test, respectively, each with the same proportion of anomalies as the whole dataset.

\begin{itemize}
\item \textbf{BlogCatalog} \cite{tang2009relational} is an online social network for users to share blogs with the public. The network structure represents the follower-followee relations among users and node attributes are a list of tags describing the users.

\item \textbf{ACM} \cite{tang2008arnetminer} is a widely used scientific citation network where nodes depict publications and edges represent the citation relations. Node attributes are extracted from the publication content using bag-of-words.

\item \textbf{Flickr} \cite{tang2009relational} is an online image-sharing network. The network structure is organized in a similar way to BlogCatalog while node attributes are generated using tags that reflect user interests.
\end{itemize}

\begin{table}[]
\centering
\caption{Dataset Statistics}
\begin{tabular}{ccccc}
    \hline
    \textbf{Dataset}  & \#Nodes & \#Edges & \#Attributes & \#Anomalies \\ \hline
    BlogCatalog & 5,196 & 172,759 & 8,189 & 298 \\ 
    ACM & 16,484 & 74,073 & 8,337 & 597 \\ 
    Flickr & 7,575 & 241,277 & 12,047 & 445 \\ \hline
    \multicolumn{5}{c}{\footnotesize Note: \#Edge counts edges excluding self-loops.}
    \end{tabular}
    \label{tb:dataset_statistics}
\end{table}

\begin{table*}
\caption{Macro- F1-score, Precision, and Recall on three datasets. (Best in bold)}
\begin{center}
\begin{tabular*}{\hsize}{@{}@{\extracolsep{\fill}}cccccccccc@{}}\toprule
\multirow{2}{*}{\textbf{Method}} & \multicolumn{3}{c}{\textbf{BlogCatalog}} & \multicolumn{3}{c}{\textbf{ACM}} & \multicolumn{3}{c}{\textbf{Flickr}}\\
\cmidrule(l){2-4} \cmidrule(l){5-7} \cmidrule(l){8-10}
&{\textbf{F1-score}} & {\textbf{Precision}}   &{\textbf{Recall}} & {\textbf{F1-score}} & {\textbf{Precision}}   &{\textbf{Recall}} &{\textbf{F1-score}} & {\textbf{Precision}}   &{\textbf{Recall}}  \\  
\hline
GCN-Detector & 0.5160 & 0.6126 & 0.5162 & 0.6434 & 0.7649 & 0.6028 & 0.7263 & \textbf{0.9147} & 0.6643
\\

GAT-Detector & 0.4931 & 0.6603 & 0.5036 & 0.7359 & \textbf{0.8084} & 0.6931 & 0.6853 & 0.7888 & 0.6464
\\

% GAT-Detector & 0.4852  & 0.4713 & 0.5000 & 0.7066  &  0.7632 & 0.6713 & 0.7001 & 0.7634 & 0.6640 
% \\

GraphSAGE-Detector & 0.6458  & 0.7810 & 0.6041 & 0.5667 & 0.7234 & 0.5434 & 0.6478 & 0.7847 & 0.6055
\\

GeniePath-Detector & 0.4852 & 0.4713  & 0.5000 & 0.4908 & 0.4819  & 0.5000  & 0.4849  & 0.4706 & 0.5000
\\

FdGars    & 0.4711  & 0.5156 & 0.5611 & 0.4108  & 0.5142  & 0.5998 & 0.6002 & 0.5814  & 0.6619
\\

DAGAD-GCN(\textbf{Ours})  & \textbf{0.8400} & \textbf{0.8480} & 0.8340 & \textbf{0.8300} &  0.7920 & \textbf{0.8840} & 0.8140  &  0.7840 & 0.8560 
\\

DAGAD-GAT(\textbf{Ours})  & 0.8080  &  0.7740 & \textbf{0.8540} & 0.7820  &  0.7460 & 0.8280 & \textbf{0.8420} &  0.8220 & \textbf{0.8600}
\\
\bottomrule

\end{tabular*}
\end{center}
\label{tb:result_semi}
\end{table*}

% \begin{table*}
% \caption{Macro- F1-score, Precision, and Recall on three datasets. (Best in bold)}
% \begin{center}
% \begin{tabular*}{\hsize}{@{}@{\extracolsep{\fill}}cccccccccc@{}}\toprule
% \multirow{2}{*}{\textbf{Method}} & \multicolumn{3}{c}{\textbf{BlogCatalog}} & \multicolumn{3}{c}{\textbf{ACM}} & \multicolumn{3}{c}{\textbf{Flickr}}\\
% \cmidrule(l){2-4} \cmidrule(l){5-7} \cmidrule(l){8-10}
% &{\textbf{F1-score}} & {\textbf{Precision}}   &{\textbf{Recall}} & {\textbf{F1-score}} & {\textbf{Precision}}   &{\textbf{Recall}} &{\textbf{F1-score}} & {\textbf{Precision}}   &{\textbf{Recall}}  \\  
% \hline
% GCN-Detector & 0.1686 & 0.5968 & 0.0996 & 0.3208 & 0.6114 & 0.2180 & 0.4252  & 0.6801  & 0.3098 \\

% GAT-Detector & 0.2803 & 0.4173 & 0.2159 & 0.4346 & 0.5376 & 0.3657 & 0.3970  & 0.6112 & 0.2952 \\

% GraphSAGE-Detector & 0.6458 & 0.7810 & 0.6041 & 0.5667 & 0.7234 & 0.5434 & 0.6478 & 0.7847 & 0.6055 \\

% GeniePath-Detector & 0.4852 & 0.4713 & 0.5000 & 0.4908 & 0.4819 & 0.5000 & 0.4849 & 0.4706 & 0.5000 \\

% FdGars & 0.4711 & 0.5156 & 0.5611 & 0.4108 & 0.5142  & 0.5998 & 0.6002 & 0.5814 & 0.6619 \\

% DAGAD-GCN(\textbf{Ours})  & \textbf{0.8400} & \textbf{0.8480} & 0.8340 & \textbf{0.8300} &  \textbf{0.7920} & \textbf{0.8840} & 0.8140  &  0.7840 & 0.8560  \\

% DAGAD-GAT(\textbf{Ours})  & 0.8080  &  0.7740 & \textbf{0.8540} & 0.7820  &  0.7460 & 0.8280 & \textbf{0.8420} &  \textbf{0.8220} & \textbf{0.8600} \\
% \bottomrule

% \end{tabular*}
% \end{center}
% \label{tb:result_semi}
% \end{table*}

\subsubsection{Baselines}
We test our proposed framework by comparison with 10 representative and up-to-date deep graph learning-based anomaly detectors as follows:

\begin{itemize}
\item \textbf{GCN-Detector} \cite{kipf2017gcn} is an anomaly detector composed of a 2-layered graph convolutional network encoder and one fully connected layer that assigns labels to nodes directly.

\item \textbf{GAT-Detector} \cite{velivckovic2018graph} adopts a 2-layered graph attention neural network for guiding the information aggregation process in graph representation learning, and similar to GCN-Detector, nodes are then identified as anomalous or normal using their representations.

\item \textbf{GraphSAGE-Detector} \cite{hamilton2017inductive} samples neighboring nodes for generating node representations and inductively spots anomalous nodes.

\item \textbf{GeniePath-Detector}\footnote{ \url{https://github.com/pygod-team/pygod} \label{pygod}} \cite{liu2019geniepath} takes a new neighbor information filtering process in GCN by exploring informative sub-graphs and paths for graph anomaly detection.

\item \textbf{FdGars}\textsuperscript{\ref{pygod}} \cite{wang2019fdgars} aims at detecting anomalies in online review networks by modeling their behavioral features and relationships from the review logs.

\item \textbf{DONE}\textsuperscript{\ref{pygod}} \cite{Bandyopadhyay2020outlier}is an Autoencoder-based anomaly detection model that identifies anomalous nodes with regard to their high structure and attribute reconstruction error. Each node's anomaly score is automatically learned through the learning process and top-$K$ nodes with higher scores are identified as anomalies.

\item \textbf{AdONE}\textsuperscript{\ref{pygod}} \cite{Bandyopadhyay2020outlier} further extends DONE under a generative-adversarial neural network framework to learn the anomaly scores and depicts anomalies in the same way as DONE.

\item \textbf{DOMINANT}\footnote{ \url{https://github.com/kaize0409/GCN_AnomalyDetection_pytorch} } \cite{Ding2019deep} adopts GCN for encoding the graph and reconstructs node attributes and graph topology using two different decoders. Each node is assigned an anomaly score with regard to its reconstruction error. 

\item \textbf{AnomalyDAE}\footnote{ \url{https://github.com/haoyfan/AnomalyDAE} } \cite{fan2020anomalydae} employs an encoder with graph attention layers for encoding the graph and a fully connected neural network for encoding node attributes. Similar to DOMINANT, anomaly scores are calculated based on the reconstruction loss. 

\item \textbf{OCGNN}\footnote{ \url{https://github.com/WangXuhongCN/myGNN} } \cite{wang2021ocgnn} adopts hypersphere learning to the decision boundaries between anomalies and normal nodes. The one-class graph neural network proposed in this work is trained using normal data, and it identifies anomalies as nodes that are precluded from the learned hypersphere. 
\end{itemize}

\subsubsection{Experimental Setting}

In our experiments, we use the published implementations of baselines and set the hyperparameter values as provided in their original papers, if available. Specifically, for GCN-Detector, GAT-Detector, GraphSAGE-Detector, DOMINANT, AnomalyDAE, and OCGNN, we implement them using the code published by their authors, and the other baselines are implemented using code provided by an open-source library\textsuperscript{\ref{pygod}} for graph anomaly detection.

Our model is implemented in Pytorch \cite{paszke2019pytorch}. We set the embedding dimensions of the GNN layer and MLP's hidden and output layers to 64, 32, and 2, respectively. $\alpha$ and $\beta$ terms in the overall loss function are set to 1.5 and 0.5, and $q$ in the GCE loss term is set to 0.7. We employ the Adam optimizer \cite{kingma2015adam} with a learning rate set to 0.005 and implement two variants of our framework based on the GCN encoders and the GAT encoders, namely DAGAD-GCN and DAGAD-GAT, to validate DAGAD's effectiveness. For DAGAD-GAT, we adopt a GAT layer with 8-head attention. The code for our framework can be accessed online\footnote{ \url{https://github.com/FanzhenLiu/DAGAD} }.

\begin{figure*}[!t]
    \centerline{\includegraphics[width=1\textwidth]{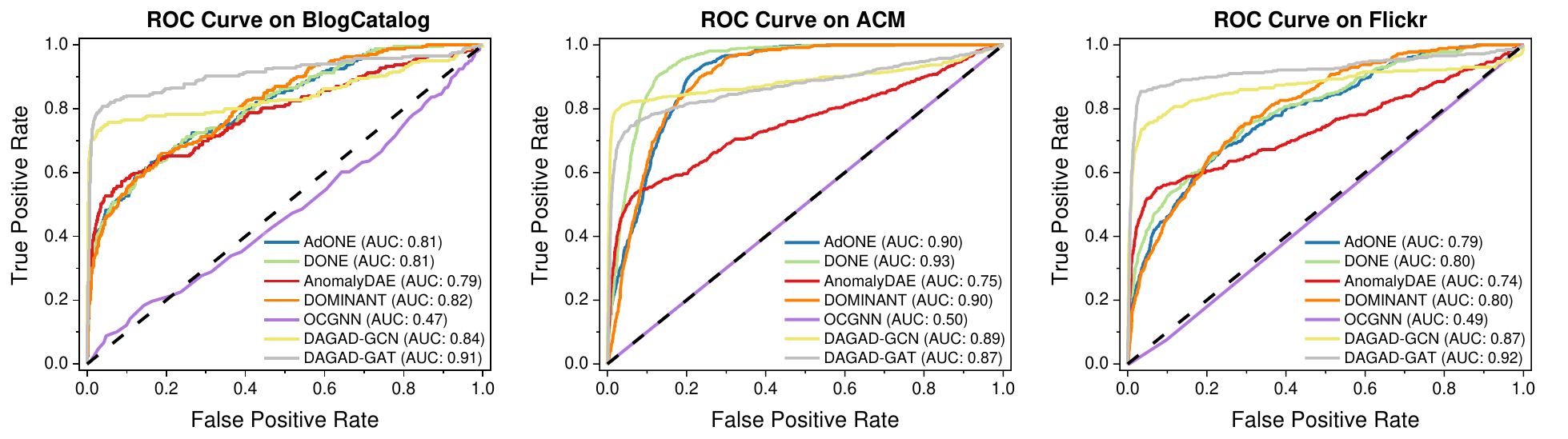}}
    \caption{ROC Curves and AUC scores of different models on BlogCatalog, ACM, and Flickr.}
    \label{pic:roc}
\end{figure*}

\subsubsection{Evaluation Metrics}
\label{metrics}

We evaluate the detection performance with five popular metrics \cite{Ding2019deep, wang2021ocgnn, wang2019fdgars}, i.e., Macro-Precision, Macro-Recall, Macro-F1-score, ROC Curve, and AUC score, since the macro metrics can reflect the performance of anomaly detectors on imbalanced data without overly underestimating the minority (anomalous) class.
Specifically, Macro-Precision is the unweighted mean of the proportions of the true anomalies in all detected anomalies and the true normal nodes in all identified normal ones, calculated by
\begin{equation}
\small
\label{precision}
\begin{aligned}
    Precision_{macro} & = \frac{P_a + P_n}{2} \\
    & = \frac{tp_a}{2(tp_a + fp_a)} + \frac{tp_n}{2(tp_n + fp_n)},
\end{aligned}
\end{equation}
where $P_a$ and $P_n$ measure the precision for the anomalous and normal classes, respectively. $tp_a$ counts the true positives of detected anomalies and $fp_a$ counts the false positives of anomalies, while $tp_n$ and $fp_n$ are those for the normal class. 

Similarly, Macro-Recall indicates the unweighted average of the proportions of the detected true anomalies in ground truth anomalies and the detected true normal nodes in ground truth normal ones. Macro-recall is valued by
\begin{equation}
\small
\label{recall}
\begin{aligned}
    Recall_{macro} & = \frac{R_a + R_n}{2} \\
    & = \frac{tp_a}{2(tp_a + fn_a)} + \frac{tp_n}{2(tp_n + fn_n)}.
\end{aligned}
\end{equation}
where $R_a$ and $R_n$ measure the recall of the anomalous and normal classes, respectively. $fn_a$ counts the false negatives of detected anomalies and $fn_n$ counts the false negatives of detected normal objects.

F1-score is a compromise between recall and precision, and Macro-F1-score is the unweighted average of the F1-scores for the two classes (i.e., anomalous and normal), quantified by:
\begin{equation}
\small
\label{f1}
\begin{aligned}
    F1_{macro} = \frac{F1_a + F1_n}{2} = \frac{P_a \cdot R_a}{P_a+R_a} + \frac{P_n \cdot R_n}{P_n+R_n}.
\end{aligned}
\end{equation}

Different from Precision, Recall, and F1-score that directly count different rates based on true positives, false positives, and false negatives, ROC and AUC measure a detector's ability in distinguishing anomalies and normal nodes by adaptively measuring the detector's true positive rate and false positive rate at various thresholds, which are widely employed to validate unsupervised detectors' performance.

\subsection{Detection Performance (Q1)}
We test the anomaly detection capabilities of our proposed framework and baselines with the widely adopted evaluation metrics introduced in Section \ref{metrics}. Specifically, for baselines trained under the guide of limited labeled samples, i.e., FdGars, GCN-, GAT-, GraphSAGE-, and GeniePath-Detector, we assess their performance concerning their macro precision, recall, and F1-scores, while for other unsupervised models, we compare the ROC curve and the AUC score. Moreover, we report the average metric values over ten runs for our models and baselines.

\subsubsection{Comparison with Semi-supervised Detectors}

The Precision, Recall, and F1-score of all baseline models and two variants of DAGAD are reported in Table~\ref{tb:result_semi}. In terms of the F1-score which shows the effectiveness of anomaly detection in the data imbalanced scenario, DAGAD employing either a one-layer GCN or GAT as GNN encoders achieves improved detection performance, that is, higher than the other five baseline methods on all datasets. Such superior detection performance benefits from our carefully designed modules. As the loss function balances the importance of training nodes in both normal and anomalous classes, our model can learn a classifier that avoids overfitting to the normal class with more training instances. Therefore, through training, assisted by a classifier trying to learn much information from the anomalous samples, GNN encoders and another classifier can eventually learn the boundary between anomalous and normal nodes and identify anomalies accurately.

The Recall is another vital measurement of a model's ability to identify all anomalies in the test data. The higher the score is, the more real anomalies are identified correctly by a model. In our experiments, both DAGAD-GCN and DAGAD-GAT achieve significantly higher recall scores than baselines, that is, nearly at least 19\% higher than any baseline, indicating that our model is superior in depicting true anomalies with much fewer anomalies being misidentified as normal.
Furthermore, as a higher score of precision means that fewer normal nodes are pinpointed as anomalous, we get the observations that the two variants of DAGAD have a better capability of depicting anomalies with lower false positive rates on BlogCatalog. GCN-Detector and GAT-Detector achieve the highest precision on Flickr and ACM, respectively, at the price of detecting real anomalies much fewer than DAGAD.

% baselines' performance analysis, overall comparison.
In summary, GCN-, GAT-, GraphSAGE-, GeniePath-Detector, and FdGars all suffer from lower F1-score and Recall, because they lack the ability to capture the distribution or patterns of true anomalies leveraging a very limited number of labeled anomalous samples in imbalanced data scenarios. Therefore, we propose DAGAD to overcome these limitations and the experimental results verify the efficacy of DAGAD regarding the most commonly used metrics. We also conduct ablation analysis in Section~\ref{sec:abl} to further explore the efficacy of the two key modules explained in Sections~\ref{module2} and \ref{module3}.

\subsubsection{Comparison with Unsupervised Detectors}

\begin{table}[]\centering
\caption{AUC score on three datasets. (Best in bold)}
  \begin{tabular}{cccc}
    \toprule
    \textbf{Method} & \textbf{BlogCatalog} & \textbf{ACM} & \textbf{Flickr} \\  
    \midrule
		DONE & 0.8108 & \textbf{0.9345} & 0.7976
		\\

		ADONE & 0.8109 & 0.9064  & 0.7929
		\\

		DOMINANT & 0.8163 & 0.8965 & 0.8029
		\\

		AnomalyDAE & 0.7860 & 0.7530 &0.7398
		\\

		OCGNN & 0.5550  & 0.5000 & 0.4891
		\\
		DAGAD-GCN(\textbf{Ours}) & \textbf{0.8364} & 0.8923  & 0.8787
		\\
		DAGAD-GAT(\textbf{Ours}) & 0.8325  & 0.8603 & \textbf{0.8955} 
		\\
    \bottomrule
  \end{tabular}
  \label{tb:result_unsup}
\end{table}

Another five most-up-to-date graph anomaly detection models are involved in comparison with DAGAD on the three datasets, and the results with regard to the ROC curve and the average AUC score are reported in Fig.~\ref{pic:roc} and Table~\ref{tb:result_unsup}, respectively.

As observed in Fig.~\ref{pic:roc}, all models except OCGNN could identify anomalies and normal nodes under different false positive rate thresholds. DAGAD-GCN and DAGAD-GAT achieve better performance than the baselines on BlogCatalog and Flickr under evaluation of the ROC curve and the AUC score. Both get remarkably higher true positive rates, e.g., nearly 0.78 and 0.86 on BlogCatalog, 0.83 and 0.90 on Flickr under the false positive rate around 0.2. Also, they achieve higher average AUC scores than any other baseline on the two datasets, as observed from Table~\ref{tb:result_unsup}.
On the ACM dataset, the two variants of our model could get relatively higher true positive rates when the false positive rate is under 0.07. The comparable AUC score shows that more comprehensive information from the structure, attribute and combined anomalies defined in DONE \cite{Bandyopadhyay2020outlier} can contribute a higher AUC score on this dataset, and this would be explored in our future work.

\subsection{Ablation Study (Q2, Q3)}
\label{sec:abl}

\begin{table}[]
\centering
\caption{Ablation Test. (best in bold)}
% \resizebox{0.48\textwidth}{!}{
\begin{tabular}{cl|ccc}
\hline
\textbf{Dataset} &  \textbf{Variants} & \textbf{F1-score} & \textbf{Precision} & \textbf{Recall}  \\ \hline
\multirow{6}{*}{\textbf{BlogCatalog}} & DAGAD-GCN\rlap{\textsuperscript{--AUG}}\textsubscript{--IMB} & 0.1686  & 0.5968  & 0.0996  \\
                                      & DAGAD-GCN\textsubscript{--IMB} & 0.2269  & 0.8085  & 0.1347  \\
                                      & DAGAD-GCN & \textbf{0.8400} & \textbf{0.8480}  & \textbf{0.8340}   \\
                                      & DAGAD-GAT\rlap{\textsuperscript{--AUG}}\textsubscript{--IMB}  & 0.2803  & 0.4173 & 0.2159   \\
                                      & DAGAD-GAT\textsubscript{--IMB}  & 0.4495  & 0.6315  & 0.3498  \\
                                      & DAGAD-GAT & \textbf{0.8080} & \textbf{0.7740} & \textbf{0.8540}    \\ \hline
                                      
\multirow{6}{*}{\textbf{ACM}}         & DAGAD-GCN\rlap{\textsuperscript{--AUG}}\textsubscript{--IMB}  & 0.3208  & 0.6114 & 0.2180   \\
                                      & DAGAD-GCN\textsubscript{--IMB} & 0.4498  & 0.7526  & 0.3259   \\
                                      & DAGAD-GCN & \textbf{0.8300} & \textbf{0.7920} & \textbf{0.8840}     \\
                                      & DAGAD-GAT\rlap{\textsuperscript{--AUG}}\textsubscript{--IMB}  & 0.4346   & 0.5376  & 0.3657 \\
                                      & DAGAD-GAT\textsubscript{--IMB}  & 0.5412  & 0.7444 & 0.4259 \\
                                      & DAGAD-GAT & \textbf{0.7820}  & \textbf{0.7460} & \textbf{0.8280}   \\ \hline
                                      
\multirow{6}{*}{\textbf{Flickr}}      & DAGAD-GCN\rlap{\textsuperscript{--AUG}}\textsubscript{--IMB} & 0.4252  & 0.6801  & 0.3098   \\
                                      & DAGAD-GCN\textsubscript{--IMB} & 0.5136 & 0.7840 & 0.3725   \\
                                      & DAGAD-GCN & \textbf{0.8140} & \textbf{0.8271} & \textbf{0.8560}  \\
                                      & DAGAD-GAT\rlap{\textsuperscript{--AUG}}\textsubscript{--IMB}  & 0.3970  & 0.6112  & 0.2952 \\                                      
                                      & DAGAD-GAT\textsubscript{--IMB} & 0.5949  & 0.7873  & 0.4796  \\
                                      & DAGAD-GAT & \textbf{0.8420} & \textbf{0.8220} & \textbf{0.8600}  \\\hline

\end{tabular}
    \label{tb:ablation}
    %}
\end{table}

Recalling the two functional modules, namely the data augmentation module and the imbalance-tailored learning module specially designed for handling the anomalous sample scarcity and class imbalance challenges associated with anomaly detection, to validate their efficacy, we conduct extensive ablation tests on the three datasets with another 4 variants involved as follows:

\begin{itemize}
    \item DAGAD-GCN\rlap{\textsuperscript{--AUG}}\textsubscript{--IMB} excludes the data augmentation module and the imbalanced-tailored learning module. It employs one GCN layer and an MLP with two fully connected neural network layers for architecture, and adopts the standard instance-wise CE loss for training.
    \item DAGAD-GAT\rlap{\textsuperscript{--AUG}}\textsubscript{--IMB} is similar to DAGAD-GCN\rlap{\textsuperscript{--AUG}}\textsubscript{--IMB}, and the only difference is that it employs a graph attention layer instead of a GCN layer.
    \item DAGAD-GCN\textsubscript{--IMB} equipped with the data augmentation module has the same neural network architecture as DAGAD-GCN, but it is trained without using the imbalance-tailored learning module.
    \item DAGAD-GAT\textsubscript{--IMB} based on the architecture of DAGAD-GAT adopts the same training strategy as DAGAD-GCN\textsubscript{--IMB}.
    % 
    % \item GCN-MLP base model (GCN base). This variant model employs one GCN layer and an MLP with two fully connected neural network layers. The parameters are fine-tuned using the conventional CE loss.
    % \item GAT-MLP base model (GAT base). Similar to the GCN base, this base model employs one graph attention layer instead of a GCN layer.
    % \item GCN base with augmentation (GCN base + AUG). This model equipped with the data augmentation module has the same neural network architecture as DAGAD-GCN, but it is trained without using the imbalance-tailored learning module.
    % \item GAT base with augmentation (GAT base + AUG). Under the DAGAD-GAT framework, this variant adopts the same training strategy as GCN base + AUG. 
\end{itemize}

% To be noted, if we stack the GCN/GAT base with data augmentation module and imbalance-tailored learning module, the model will be DAGAD-GCN/GAT (the ``+IMB" line in Table~\ref{tb:ablation}). 
As shown in Table~\ref{tb:ablation}, when applying additional learning modules to vanilla GCN- and GAT-based models, the anomaly detection performance is improved dramatically regarding the F1-score, Precision, and Recall. Specifically, with the imbalance-tailored learning module adopted to guide the learning process, the F1-score and the Recall of DAGAD increase by at least 41\% and 79\% on the three datasets, respectively. Additionally, the data augmentation module helps to achieve a precision increase of over 35\% on BlogCatalog, 23\% on ACM, and 15\% on Flickr.
% Specifically, after adopting the imbalance-tailored learning module to guide the learning process, the F1-score increases by over 34\%, 40\%, and 20\% on BlogCatalog, ACM, and Flickr, respectively. 
To this end, we can safely conclude that both the data augmentation module and the imbalance-tailored learning module effectively facilitate anomaly detection performance.

\subsection{Sensitivity to Hyper-parameters (Q4)} \label{sec:sensitivity}

% \begin{figure}[]
%     \centerline{\includegraphics[width=0.48\textwidth]{pic/sensitivity.pdf}}
%     \caption{Hyper-parameter sensitivity analysis.}
%     \label{pic:sens}
% \end{figure}

% \begin{figure}[]
% 	\centering
% 	\subfigure[]{
% 		\begin{minipage}{4.25cm}
% 			\centering
% 			\includegraphics[width=1\textwidth]{pic/sensitivity-precision.pdf}
% 		\end{minipage}
% 	}\hspace{-0.3cm}
% 	%
% 	\subfigure[]{
% 		\begin{minipage}{4.25cm}
% 			\centering
% 			\includegraphics[width=1\textwidth]{pic/sensitivity-recall.pdf}
% 		\end{minipage}
% 	}
% 	\subfigure[]{
% 		\begin{minipage}{4.25cm}
% 			\centering
% 			\includegraphics[width=1\textwidth]{pic/sensitivity-F1.pdf}
% 		\end{minipage}
% 	}\hspace{-0.3cm}
% 	%
% 	\subfigure[]{
% 		\begin{minipage}{4.25cm}
% 			\centering
% 			\includegraphics[width=1\textwidth]{pic/sensitivity-AUC.pdf}
% 		\end{minipage}
% 	}
% 	\caption{DAGAD-GCN Hyper-parameter sensitivity analysis with four metrics: (a) Precision, (b) Recall, (c) F1-score, and (d) AUC.}
% 	\label{pic:gcn_alpha_sens}
% \end{figure}

\begin{figure}[]
    \centerline{\includegraphics[width=0.45\textwidth]{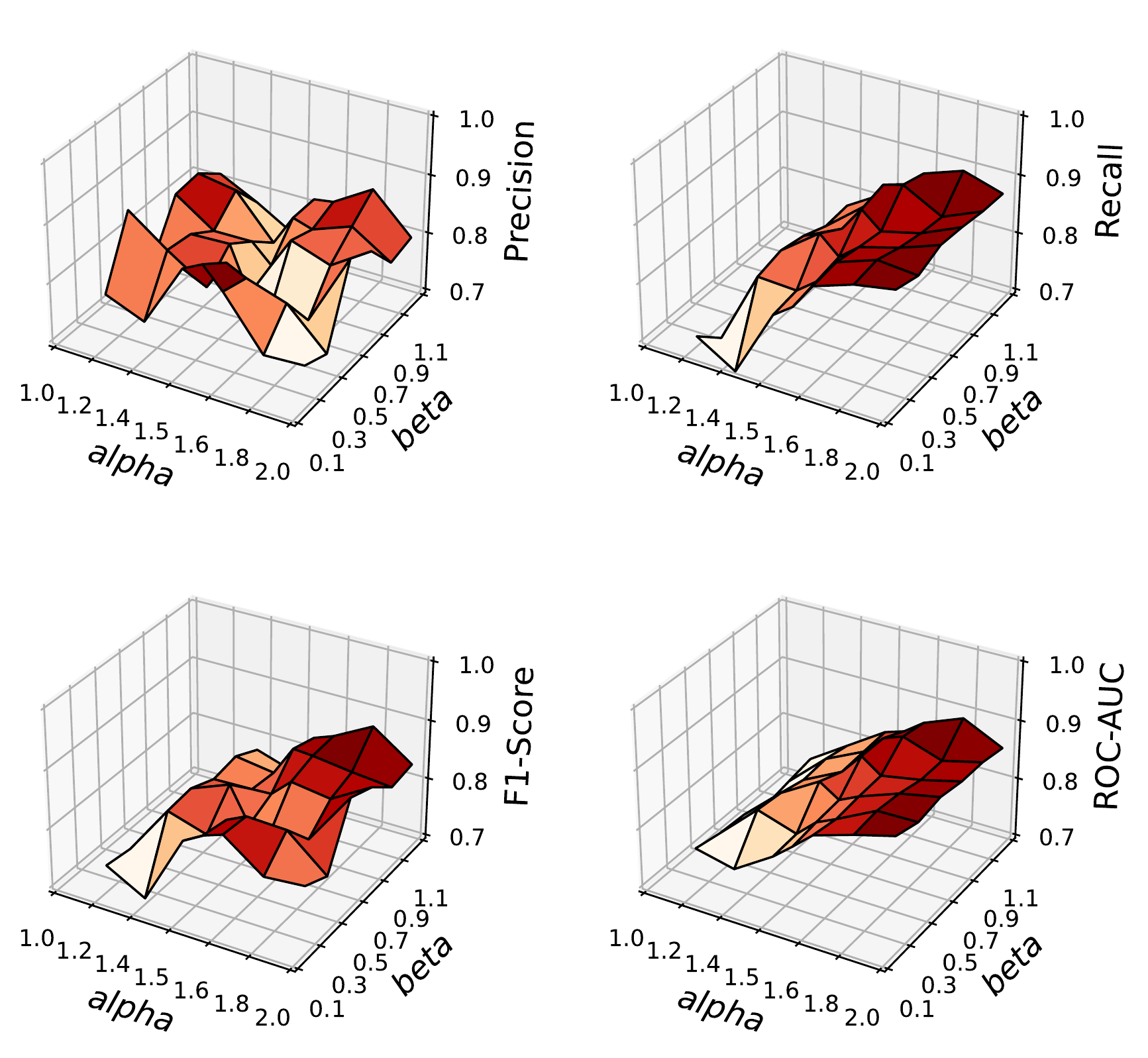}}
    \caption{Parameter sensitivity analysis of DAGAD-GCN on BlogCatalog by four metrics: Precision, Recall, F1-score, and AUC score.}
    \label{pic:gcn_alpha_sens}
\end{figure}

\begin{figure}[]
    \centerline{\includegraphics[width=0.45\textwidth]{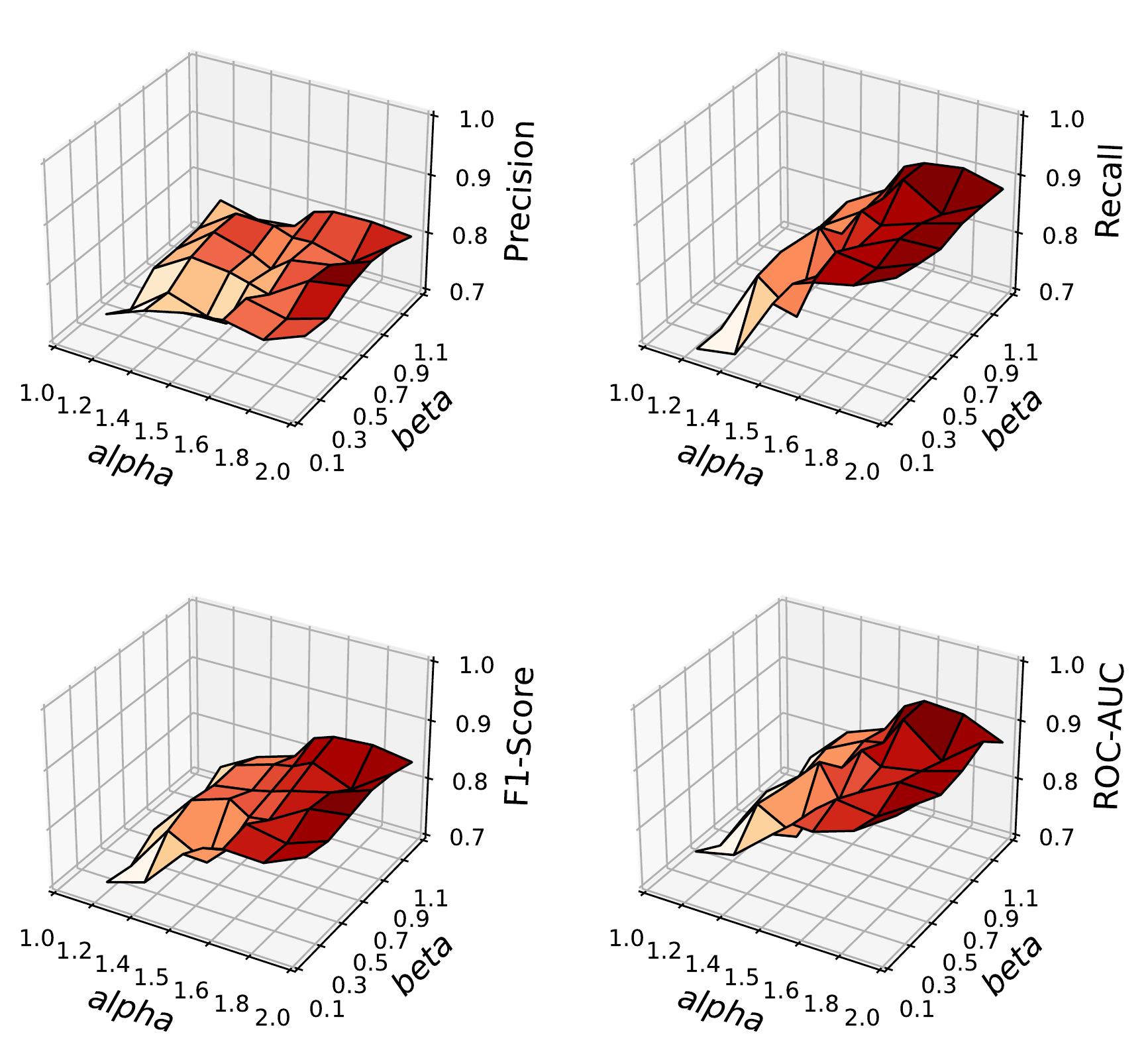}}
    \caption{Parameter sensitivity analysis of DAGAD-GAT on BlogCatalog by four metrics: Precision, Recall, F1-score, and AUC score.}
    \label{pic:gat_alpha_sens}
\end{figure}

% \begin{figure}[]
% 	\centering
% 	\subfigure[]{
% 		\begin{minipage}{4.25cm}
% 			\centering
% 			\includegraphics[width=1\textwidth]{pic/GAT-precision.pdf}
% 		\end{minipage}
% 	}\hspace{-0.3cm}
% 	%
% 	\subfigure[]{
% 		\begin{minipage}{4.25cm}
% 			\centering
% 			\includegraphics[width=1\textwidth]{pic/GAT-recall.pdf}
% 		\end{minipage}
% 	}
% 	\subfigure[]{
% 		\begin{minipage}{4.25cm}
% 			\centering
% 			\includegraphics[width=1\textwidth]{pic/GAT-F1.pdf}
% 		\end{minipage}
% 	}\hspace{-0.3cm}
% 	%
% 	\subfigure[]{
% 		\begin{minipage}{4.25cm}
% 			\centering
% 			\includegraphics[width=1\textwidth]{pic/GAT-AUC.pdf}
% 		\end{minipage}
% 	}
% 	\caption{DAGAD-GAT Hyper-parameter sensitivity analysis with four metrics: (a) Precision, (b) Recall, (c) F1-score, and (d) AUC.}
% 	\label{pic:gat_alpha_sens}
% \end{figure}

%\subsubsection{$\alpha$ and $\beta$ sensitivity study}

The two regularization terms $\alpha$ and $\beta$ in DAGAD's overall loss function as shown in Eq.~\eqref{overall_loss_new} are critical in balancing the weights of training losses introduced by $C_a$, $C_b$, and the data augmentation module. Hence, we evaluate our model's sensitivity to these two terms by comparing Precision, Recall, F1-score, and AUC score under different settings. Specifically, we run DAGAD-GCN and DAGAD-GAT on BlogCatalog dataset for $\alpha \in \{1.0, 1.2, 1.4, 1.5, 1.6, 1.8, 2.0\}$ and $\beta \in \{0.1, 0.3, 0.5, 0.7, 0.9, 1.1\}$ and report the results in Figs.~\ref{pic:gcn_alpha_sens} and \ref{pic:gat_alpha_sens}. It can be observed that the better values for $\alpha$ and $\beta$ are around 1.6 and 0.9, respectively.

\section{Conclusions}
\label{concl}

In this work, we discuss two paramount challenges associated with graph anomaly detection, i.e., anomalous sample scarcity and class imbalance. Both challenges introduce defective results to current works in this field because a clear boundary between anomalies and normal objects cannot be effectively learned with insufficient knowledge about anomalies. To alleviate these issues, we devise a data augmentation-based graph anomaly detection framework, DAGAD. 
% and conduct extensive experiments to show its superiority by comparison with 10 up-to-date graph anomaly detectors.
% Specifically, we develop a data augmentation module enriching the training set based on comprehensive representations and an imbalance-tailored learning module balancing the contribution of the majority and minority classes to the detection process. Resolving anomaly detection as a binary classification task, DAGAD employs a GNN-based information fusion module to learn representations, serving the above two modules. 
With three widely used datasets, extensive experiments present a comprehensive assessment of detectors regarding Macro-Precision, Macro-Recall, Macro-F1-score, ROC curve, and AUC score. The results demonstrate that DAGAD outperforms 10 up-to-date baselines, followed by an ablation study validating the power of our proposed modules. This work is expected to provide a promising solution for practical applications with class-imbalanced, limited labeled data.

\section*{Acknowledgment}
This work was supported by the ARC DECRA Project (No. DE200100964), the ARC Discovery Project (No. DP200102298), and the NSFC (No. 61872360). 
% The first two authors contributed equally to this work.

\bibliographystyle{IEEEtran}
\bibliography{ref}

\begin{thebibliography}{10}
\providecommand{\url}[1]{#1}
\csname url@samestyle\endcsname
\providecommand{\newblock}{\relax}
\providecommand{\bibinfo}[2]{#2}
\providecommand{\BIBentrySTDinterwordspacing}{\spaceskip=0pt\relax}
\providecommand{\BIBentryALTinterwordstretchfactor}{4}
\providecommand{\BIBentryALTinterwordspacing}{\spaceskip=\fontdimen2\font plus
\BIBentryALTinterwordstretchfactor\fontdimen3\font minus
  \fontdimen4\font\relax}
\providecommand{\BIBforeignlanguage}[2]{{%
\expandafter\ifx\csname l@#1\endcsname\relax
\typeout{** WARNING: IEEEtran.bst: No hyphenation pattern has been}%
\typeout{** loaded for the language `#1'. Using the pattern for}%
\typeout{** the default language instead.}%
\else
\language=\csname l@#1\endcsname
\fi
#2}}
\providecommand{\BIBdecl}{\relax}
\BIBdecl

\bibitem{chandola2009anomaly}
V.~Chandola, A.~Banerjee, and V.~Kumar, ``Anomaly detection: A survey,''
  \emph{ACM Comput. Surv.}, vol.~41, no. 3, Art. no. 15, 2009.

\bibitem{pang2021deep}
G.~Pang, C.~Shen, L.~Cao, and A.~V.~D. Hengel, ``Deep learning for anomaly
  detection: A review,'' \emph{ACM Comput. Surv.}, vol.~54, no. 2, Art. no. 38,
  2021.

\bibitem{shu2017fake}
K.~Shu, A.~Sliva, S.~Wang, J.~Tang, and H.~Liu, ``Fake news detection on social
  media: A data mining perspective,'' \emph{SIGKDD Explor. Newsl.}, vol.~19,
  no.~1, pp. 22--36, 2017.

\bibitem{akoglu2015graph}
L.~Akoglu, H.~Tong, and D.~Koutra, ``Graph based anomaly detection and
  description: A survey,'' \emph{Data Min. Knowl. Disc.}, pp. 626--688, 2015.

\bibitem{miller2014twitter}
Z.~Miller, B.~Dickinson, W.~Deitrick, W.~Hu, and A.~H. Wang, ``Twitter spammer
  detection using data stream clustering,'' \emph{Inf. Sci.}, vol. 260, pp.
  64--73, 2014.

\bibitem{chakrabarti2006graph}
D.~Chakrabarti and C.~Faloutsos, ``Graph mining: Laws, generators, and
  algorithms,'' \emph{ACM Comput. Surv.}, vol.~38, no.~1, p. 2–es, 2006.

\bibitem{hamilton2017representation}
W.~Hamilton, R.~Ying, and J.~Leskovec, ``Representation learning on graphs:
  Methods and applications,'' \emph{IEEE Data Eng. Bull.}, vol.~40, no.~3, pp.
  52--74, 2017.

\bibitem{ma2021survey}
X.~Ma, J.~Wu, S.~Xue, J.~Yang, C.~Zhou, Q.~Z. Sheng, H.~Xiong, and L.~Akoglu,
  ``A comprehensive survey on graph anomaly detection with deep learning,''
  \emph{IEEE Trans. Knowl. Data Eng.}, 2021.

\bibitem{zhang2021fraudre}
G.~Zhang, J.~Wu, J.~Yang, A.~Beheshti, S.~Xue, C.~Zhou, and Q.~Z. Sheng,
  ``{FRAUDRE}: Fraud detection dual-resistant to graph inconsistency and
  imbalance,'' in \emph{ICDM}, 2021, pp. 867--876.

\bibitem{Weber2019anti}
M.~Weber, G.~Domeniconi, J.~Chen, D.~K.~I. Weidele, T.~R. C.~Bellei, and C.~E.
  Leiserson, ``Anti-money laundering in bitcoin: Experimenting with graph
  convolutional networks for financial forensics,'' in \emph{KDD Workshop on
  Anomaly Detection in Finance}, 2019.

\bibitem{liu2022eriskcom}
F.~Liu, Z.~Li, B.~Wang, J.~Wu, J.~Yang, J.~Huang, Y.~Zhang, W.~Wang, S.~Xue,
  S.~Nepal, and Q.~Z. Sheng, ``{eRiskCom}: An e-commerce risky community
  detection platform,'' \emph{VLDB J.}, vol.~31, pp. 1085--1101, 2022.

\bibitem{chalapathy2019deep}
R.~Chalapathy and S.~Chawla, ``Deep learning for anomaly detection: {A}
  survey,'' \emph{arXiv preprint arXiv:1901.03407}, vol. abs/1901.03407, 2019.

\bibitem{Hinton2006reducing}
G.~E. Hinton and R.~R. Salakhutdinov, ``Reducing the dimensionality of data
  with neural networks,'' \emph{Science}, vol. 313, no. 5786, pp. 504--507,
  2006.

\bibitem{Wang2019a}
D.~Wang, J.~Lin, P.~Cui, Q.~Jia, Z.~Wang, Y.~Fang, Q.~Yu, J.~Zhou, S.~Yang, and
  Y.~Qi, ``A semi-supervised graph attentive network for financial fraud
  detection,'' in \emph{ICDM}, 2019, pp. 598--607.

\bibitem{Ding2021few}
K.~Ding, Q.~Zhou, H.~Tong, and H.~Liu, ``Few-shot network anomaly detection via
  cross-network meta-learning,'' in \emph{WWW}, 2021, pp. 2448--2456.

\bibitem{zhou2020graph}
J.~Zhou, G.~Cui, S.~Hu, Z.~Zhang, C.~Yang, Z.~Liu, L.~Wang, C.~Li, and M.~Sun,
  ``Graph neural networks: A review of methods and applications,'' \emph{AI
  Open}, vol.~1, pp. 57--81, 2020.

\bibitem{Wu2021GNN}
Z.~Wu, S.~Pan, F.~Chen, G.~Long, C.~Zhang, and P.~S. Yu, ``A comprehensive
  survey on graph neural networks,'' \emph{IEEE Trans. Neural Netw. Learn.
  Syst.}, vol.~32, no.~1, pp. 4--24, 2021.

\bibitem{Luo2022ComGA}
X.~Luo, J.~Wu, A.~Beheshti, J.~Yang, X.~Zhang, Y.~Wang, and S.~Xue, ``{ComGA}:
  Community-aware attributed graph anomaly detection,'' in \emph{WSDM}, 2022,
  pp. 657--665.

\bibitem{Bandyopadhyay2020outlier}
S.~Bandyopadhyay, L.~N, S.~V. Vivek, and M.~N. Murty, ``Outlier resistant
  unsupervised deep architectures for attributed network embedding,'' in
  \emph{WSDM}, 2020, pp. 25--33.

\bibitem{wang2021ocgnn}
X.~Wang, B.~Jin, Y.~Du, P.~Cui, Y.~Tan, and Y.~Yang, ``One-class graph neural
  networks for anomaly detection in attributed networks,'' \emph{Neural Comput.
  Appl.}, pp. 1--13, 2021.

\bibitem{Ding2019deep}
K.~Ding, J.~Li, R.~Bhanushali, and H.~Liu, ``Deep anomaly detection on
  attributed networks,'' in \emph{SDM}, 2019, pp. 594--602.

\bibitem{akoglu2010oddball}
L.~Akoglu, M.~McGlohon, and C.~Faloutsos, ``{OddBall}: Spotting anomalies in
  weighted graphs,'' in \emph{PAKDD}, 2010, pp. 410--421.

\bibitem{sharpnack2013near}
J.~L. Sharpnack, A.~Krishnamurthy, and A.~Singh, ``Near-optimal anomaly
  detection in graphs using lovasz extended scan statistic,'' in
  \emph{NeurIPS}, 2013, pp. 1959--1967.

\bibitem{li2017Radar}
J.~Li, H.~Dani, X.~Hu, and H.~Liu, ``Radar: Residual analysis for anomaly
  detection in attributed networks,'' in \emph{IJCAI}, 2017, pp. 2152--2158.

\bibitem{fan2020anomalydae}
H.~Fan, F.~Zhang, and Z.~Li, ``{AnomalyDAE}: Dual autoencoder for anomaly
  detection on attributed networks,'' in \emph{ICASSP}, 2020, pp. 5685--5689.

\bibitem{liu2019geniepath}
Z.~Liu, C.~Chen, L.~Li, J.~Zhou, X.~Li, L.~Song, and Y.~Qi, ``{GeniePath}:
  Graph neural networks with adaptive receptive paths,'' in \emph{AAAI}, 2019,
  pp. 4424--4431.

\bibitem{wang2019fdgars}
J.~Wang, R.~Wen, C.~Wu, Y.~Huang, and J.~Xiong, ``{FdGars}: Fraudster detection
  via graph convolutional networks in online app review system,'' in
  \emph{WWW}, 2019, pp. 310--316.

\bibitem{kipf2017gcn}
T.~N. Kipf and M.~Welling, ``Semi-supervised classification with graph
  convolutional networks,'' in \emph{ICLR}, 2017.

\bibitem{velivckovic2018graph}
P.~Veli{\v{c}}kovi{\'{c}}, G.~Cucurull, A.~Casanova, A.~Romero, P.~Li{\`{o}},
  and Y.~Bengio, ``Graph attention networks,'' in \emph{ICLR}, 2018.

\bibitem{hamilton2017inductive}
W.~L. Hamilton, R.~Ying, and J.~Leskovec, ``Inductive representation learning
  on large graphs,'' in \emph{NeurIPS}, 2017, pp. 1025--1035.

\bibitem{Shorten2019survey}
C.~Shorten and T.~M. Khoshgoftaar, ``A survey on image data augmentation for
  deep learning,'' \emph{J. Big Data}, vol.~5, pp. 221--232, 2016.

\bibitem{feng2021survey}
S.~Y. Feng, V.~Gangal, J.~Wei, S.~Chandar, S.~Vosoughi, T.~Mitamura, and
  E.~Hovy, ``A survey of data augmentation approaches for {NLP},'' in
  \emph{Findings of ACL-IJCNLP}, 2021, pp. 968--988.

\bibitem{lee2021learning}
J.~Lee, E.~Kim, J.~Lee, J.~Lee, and J.~Choo, ``Learning debiased representation
  via disentangled feature augmentation,'' in \emph{NeurIPS}, 2021, pp.
  25\,123--25\,133.

\bibitem{Suresh2021adversarial}
S.~Suresh, P.~Li, C.~Hao, and J.~Neville, ``Adversarial graph augmentation to
  improve graph contrastive learning,'' in \emph{NeurIPS}, 2021, pp.
  15\,920--15\,933.

\bibitem{Zhao2021data}
T.~Zhao, Y.~Liu, L.~Neves, O.~Woodford, M.~Jiang, and N.~Shah, ``Data
  augmentation for graph neural networks,'' in \emph{AAAI}, 2021, pp.
  11\,015--11\,023.

\bibitem{zhu2021graph}
Y.~Zhu, Y.~Xu, F.~Yu, Q.~Liu, S.~Wu, and L.~Wang, ``Graph contrastive learning
  with adaptive augmentation,'' in \emph{WWW}, 2021, pp. 2069--2080.

\bibitem{Park2021metropolis}
H.~Park, S.~Lee, S.~Kim, J.~Park, J.~Jeong, K.-M. Kim, J.-W. Ha, and H.~J. Kim,
  ``{Metropolis-Hastings} data augmentation for graph neural networks,'' in
  \emph{NeurIPS}, 2021, pp. 19\,010--19\,020.

\bibitem{kong2022robust}
K.~Kong, G.~Li, M.~Ding, Z.~Wu, C.~Zhu, B.~Ghanem, G.~Taylor, and T.~Goldstein,
  ``{FLAG}: Adversarial data augmentation for graph neural network,'' in
  \emph{CVPR}, 2022.

\bibitem{You2020graph}
Y.~You, T.~Chen, Y.~Sui, T.~Chen, Z.~Wang, and Y.~Shen, ``Graph contrastive
  learning with augmentations,'' in \emph{NeurIPS}, 2020, pp. 5812--5823.

\bibitem{Qiu2020gcc}
J.~Qiu, Q.~Chen, Y.~Dong, J.~Zhang, H.~Yang, M.~Ding, K.~Wang, and J.~Tang,
  ``{GCC}: Graph contrastive coding for graph neural network pre-training,'' in
  \emph{KDD}, 2020, pp. 1150--1160.

\bibitem{you2021graph}
Y.~You, T.~Chen, Y.~Shen, and Z.~Wang, ``Graph contrastive learning
  automated,'' in \emph{ICML}, 2021, pp. 12\,121--12\,132.

\bibitem{Johnson2019survey}
J.~M. Johnson and T.~M. Khoshgoftaar, ``Survey on deep learning with class
  imbalance,'' \emph{J. Big Data}, vol. 6, Art. no. 27, 2019.

\bibitem{yang2020rethink}
Y.~Yang and Z.~Xu, ``Rethinking the value of labels for improving
  class-imbalanced learning,'' in \emph{NeurIPS}, 2020, pp. 19\,290--19\,301.

\bibitem{Shi2020multi}
M.~Shi, Y.~Tang, X.~Zhu, D.~Wilson, and J.~Liu, ``Multi-class imbalanced graph
  convolutional network learning,'' in \emph{IJCAI}, 2020, pp. 2879--2885.

\bibitem{Zhao2021GraphSMOTE}
T.~Zhao, X.~Zhang, and S.~Wang, ``{GraphSMOTE}: Imbalanced node classification
  on graphs with graph neural networks,'' in \emph{WSDM}, 2021, pp. 833--841.

\bibitem{Qu2021ImGAGN}
L.~Qu, H.~Zhu, R.~Zheng, Y.~Shi, and H.~Yin, ``{ImGAGN}: Imbalanced network
  embedding via generative adversarial graph networks,'' in \emph{KDD}, 2021,
  pp. 1390--1398.

\bibitem{Goodfellow2016deep}
I.~Goodfellow, Y.~Bengio, and A.~Courville, \emph{Deep Learning}.\hskip 1em
  plus 0.5em minus 0.4em\relax Cambridge, MA, USA: MIT Press, 2016, ch. Deep
  Feedforward Networks, pp. 163--220.

\bibitem{zhang2018generalized}
Z.~Zhang and M.~Sabuncu, ``Generalized cross entropy loss for training deep
  neural networks with noisy labels,'' in \emph{NeurIPS}, 2018, pp. 8792--8802.

\bibitem{nam2020learning}
J.~Nam, H.~Cha, S.~Ahn, J.~Lee, and J.~Shin, ``Learning from failure:
  De-biasing classifier from biased classifier,'' in \emph{NeurIPS}, 2020, pp.
  20\,673--20\,684.

\bibitem{ding2019interactive}
K.~Ding, J.~Li, and H.~Liu, ``Interactive anomaly detection on attributed
  networks,'' in \emph{WSDM}, 2019, pp. 357--365.

\bibitem{tang2009relational}
L.~Tang and H.~Liu, ``Relational learning via latent social dimensions,'' in
  \emph{KDD}, 2009, pp. 817--826.

\bibitem{tang2008arnetminer}
J.~Tang, J.~Zhang, L.~Yao, J.~Li, L.~Zhang, and Z.~Su, ``{ArnetMiner}:
  Extraction and mining of academic social networks,'' in \emph{KDD}, 2008, pp.
  990--998.

\bibitem{paszke2019pytorch}
A.~Paszke, S.~Gross, F.~Massa, A.~Lerer, J.~Bradbury, G.~Chanan, T.~Killeen,
  Z.~Lin, N.~Gimelshein, L.~Antiga, A.~Desmaison, A.~Kopf, E.~Yang, Z.~DeVito,
  M.~Raison, A.~Tejani, S.~Chilamkurthy, B.~Steiner, L.~Fang, J.~Bai, and
  S.~Chintala, ``Pytorch: An imperative style, high-performance deep learning
  library,'' in \emph{NeurIPS}, 2019, pp. 8024--8035.

\bibitem{kingma2015adam}
D.~P. Kingma and J.~Ba, ``Adam: A method for stochastic optimization,'' in
  \emph{ICLR}, 2015.

\end{thebibliography}

\end{document}